\documentclass[letterpaper,10pt,conference]{ieeeconf}

\IEEEoverridecommandlockouts
\usepackage{iftex}
\ifXeTeX
  \usepackage{newtxtext}
\fi
\usepackage{amsmath,amssymb}
\usepackage{booktabs}
\usepackage{cite}
\usepackage{graphicx}
\usepackage{url}
\usepackage{makecell}
\usepackage[table]{xcolor}
\usepackage{tabularx}
\usepackage{nameref}
\usepackage{needspace}
\definecolor{oursgreen}{HTML}{E8F5E9}
\graphicspath{{figures/}}
\newcommand{\method}{\textsc{SE-LLM--OCP}}

\definecolor{paperBlue}{RGB}{44,92,130}
\definecolor{paperTeal}{RGB}{36,122,121}
\definecolor{paperOrange}{RGB}{194,116,34}
\definecolor{paperGreen}{RGB}{47,122,76}
\definecolor{paperRed}{RGB}{168,65,65}
\definecolor{paperGray}{RGB}{92,99,106}

\title{\LARGE \bf
From Semantic Decisions to Feasible Trajectories:\\
Self-Evolving LLM-Guided Optimal Control for Narrow-Space Parking}

\author{
Zhengbao Yao$^{1}$, Yuanfu Luo$^{1,2}$, Kehan Xue$^{1}$\\
$^{1}$All authors are with xLean Robotics Co., Ltd.\\
$^{2}$Corresponding author: Yuanfu Luo (e-mail: yuanfu@xlean.ai)%
}
\usepackage[ruled,vlined,linesnumbered]{algorithm2e}

\SetKwInput{KwInput}{Input}
\SetKwInput{KwOutput}{Output}
\SetKw{KwRet}{return}

\RestyleAlgo{ruled}
\begin{document} 
\ifPDFTeX
\fi
\maketitle
\thispagestyle{empty}
\pagestyle{empty}
\raggedbottom
\bstctlcite{BSTcontrol}

\begin{abstract}
Autonomous parking in nonconvex and narrow environments remains challenging. Although optimal-control methods can explicitly enforce vehicle dynamics and collision constraints, nonconvexity compromises solver robustness and can cause failures. Large language models (LLMs) exhibit strong semantic reasoning capabilities, but directly generating dense trajectories makes it difficult to guarantee physical feasibility. We introduce \method{}, a unified framework in which LLMs make high-level discrete maneuver decisions, while an optimal-control module enforces low-level vehicle dynamics and collision constraints. Online, the LLM proposes sparse maneuver plans, decomposing the parking task into a sequence of short-horizon trajectory-optimization problems. A low-level solver then sequentially solves optimal-control problems. If the solver fails, the LLM aggregates failure evidence from the solver and validation stages to guide replanning. Offline, \method{} automatically evolves a structured decision-making knowledge base from scratch, driven by accumulated online failures. We validate our proposed framework in simulation on a car-like vehicle model and on a differential-drive robot. Our experimental results show that \method{} enables safer autonomous parking in narrow scenarios and demonstrates transfer of the same maneuver representation to a different kinematic platform.
\end{abstract}

\section{Introduction}

Autonomous parking in cluttered environments is a canonical hybrid motion-planning problem. A vehicle trajectory must satisfy nonholonomic kinematics and full-footprint collision constraints, yet planner success often depends on discrete maneuver choices: the approach direction, the ordering of forward and reverse segments, and the placement of intermediate configurations. Search-based planners such as Hybrid A* (HA*) explore these choices with discrete nonholonomic motion primitives \cite{dolgov2008practical}. Optimization-based planners instead optimize continuous paths or trajectories under collision costs or constraints and, where modeled, vehicle kinematics or dynamics \cite{ratliff2009chomp,zhang2021optimization}. The two paradigms fail in different ways. Search effort depends on the resolution and coverage of the motion primitives, whereas nonlinear trajectory optimization is sensitive to initialization and the selected homotopy class. These weaknesses become pronounced in nonconvex and narrow spaces: an unsuitable maneuver structure may lead to excessive search or local optimization failure even though a feasible trajectory exists.

Large language models offer a possible source of maneuver-level structure when coupled with visual or structured scene representations. Language-grounded robotic systems have used foundation models to decompose instructions into admissible actions, ground skill selection in affordances, synthesize executable policy programs, and plan over scene-level representations \cite{huang2022zeroshot,ichter2023saycan,liang2023code,rana2023sayplan}. Autonomous-driving studies have likewise explored language-model reasoning for motion planning and coordinate-based trajectory generation \cite{mao2023gptdriver,tian2025drivevlm}. These approaches do not remove the physical‑grounding problem: a semantically plausible maneuver can fail because of terminal‑pose error, insufficient clearance, or an infeasible swept volume. This motivates assigning discrete maneuver structure to the LLM while retaining model‑based checks for continuous feasibility and converting low‑level failures into actionable replanning feedback.

\begin{figure*}[!t]
\centering
\includegraphics[width=\textwidth]{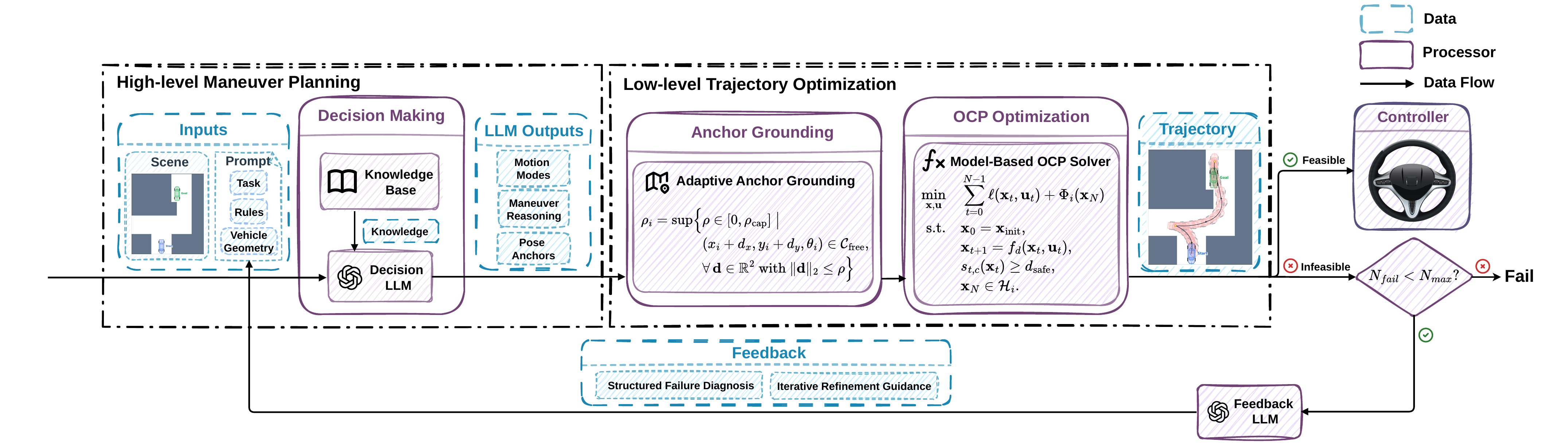}
\caption{Online hierarchical pipeline of \method{}: the
decision-making LLM turns scene and task inputs plus retrieved
parking knowledge into pose anchors and motion modes; these
define sequential segment OCPs, and accepted trajectories are
passed to the controller. A feedback LLM summarizes
failed attempts to guide maneuver revision within the attempt budget.}
\label{fig:pipeline}
\end{figure*}

This paper answers this question with \method{} (Self-Evolving LLM-Guided Optimal Control Planning), a hierarchical framework for narrow-space autonomous parking. In \method{}, an LLM serves as the maneuver-level planner: from the scene, the task, and a parking-decision knowledge base, it decides how the vehicle should maneuver---where to stage, which direction to move, and how to approach the goal---and expresses these decisions as sparse terminal poses and forward or reverse motion modes, each accompanied by a short rationale. It never generates dense trajectories or low-level controls. A deterministic optimal-control layer then turns each decision into a short-horizon optimal control problem (OCP), enforcing nonholonomic vehicle kinematics, actuation limits, and full-footprint collision avoidance. Semantic reasoning and metric optimization are thus kept in strictly separate layers, each doing what it does best.

Since LLM-proposed intermediate anchors are approximate, \method{} grounds them in clearance-adapted terminal regions with terminal penalties to favor nominal anchors, while the final goal is strictly enforced. This accommodates small semantic-to-metric errors without relaxing kinematic or collision constraints, which are independently verified by the OCP and trajectory validator. If OCP solving or trajectory validation fails, a feedback LLM aggregates solver diagnostics and validation errors with failed rationales to revise maneuvers within a bounded attempt budget. In a separate offline loop, accumulated failures drive revisions to a structured parking-decision knowledge base initialized from an empty state. Candidate updates are retained only when the revised planner passes trajectory-acceptance checks on both triggering and prior successful scenarios. The LLM parameters remain fixed; self-evolution herein refers to bounded revisions of task-level maneuver decisions and decision knowledge.

The main contributions of this work are:
\begin{itemize}
\item a hierarchical framework for narrow-space autonomous parking that pairs LLM-based maneuver reasoning with deterministic optimal control, producing physically feasible trajectories under full vehicle kinematics and collision avoidance;

\item a semantic-to-optimization grounding that absorbs the metric approximation inherent in LLM decisions, compiling them into tolerance-adapted terminal conditions for short-horizon OCPs without weakening kinematic or collision constraints;

\item a bounded, failure-conditioned self-evolution loop that converts failed attempts into maneuver-level guidance, letting the LLM recover from infeasible decisions without unbounded replanning; 

\item an offline knowledge self-evolution loop that bootstraps the parking-decision knowledge base from an initially empty state, with all entries generated automatically, and retains an update only when it yields a feasible trajectory.
\end{itemize}

We validate \method{} in simulation across confined parking tasks. Under a shared Stanley--PID controller, it achieves the highest planning and closed-loop success rates among baselines, with fewer gear changes and lower path curvature than HA*+OCP. Ablations examine the evolved knowledge base, failure feedback, and intermediate soft constraints. A differential-drive experiment further provides qualitative evidence that the same segment-level maneuver representation transfers to a different platform.

\section{Related Work}

\subsection{Traditional Planning Approaches}

Representative parking planners use search‑based methods, trajectory optimization, or their combination. Reeds‑Shepp curves provide shortest paths for a car with bounded curvature and forward‑reverse motion in obstacle‑free space~\cite{reeds1990optimal}. Hybrid A* incorporates vehicle kinematics into heuristic search and refines the resulting path through numerical optimization~\cite{dolgov2008practical}. Many optimization‑based planners generate continuous trajectories under explicit vehicle and collision constraints. Li et al. iteratively reconstruct safe corridors for parking among irregular obstacles~\cite{li2022lightweight}; Han et al. exploit differential flatness for spatiotemporal optimization~\cite{han2024spatiotemporal}; RDA accelerates collision‑constrained planning through biconvex reformulation and obstacle‑wise parallel updates~\cite{han2023rda}. Unlike methods that directly optimize continuous trajectories, \method{} uses an LLM to propose and revise sparse staging poses and motion modes, which are then grounded through sequential OCPs.

\subsection{End-to-End Parking Approaches}

End‑to‑end approaches learn parking behavior from scene observations and target information. Yang et al. investigated end‑to‑end neural parking in the CARLA simulator~\cite{yang2024e2eparking}. ParkingE2E predicts parking waypoints from surround‑view images and a target parking slot using imitation learning and an autoregressive transformer~\cite{li2024parkinge2e}. MultiPark extends learned parking to multimodal, multi‑segment prediction with queries that encode gear, longitudinal, and lateral behaviors together with goal‑ and collision‑related objectives~\cite{zheng2026multipark}. These methods learn dense or multimodal parking policies, whereas \method{} uses the language model only for sparse anchors and motion modes and delegates continuous trajectory generation and validation to model‑based OCPs.

\subsection{Grounding and Self-Evolution in LLM-Based Planning}

Grounding links language reasoning to executable robot behavior. SayCan scores candidate skills with affordances~\cite{ichter2023saycan}; AutoTAMP translates instructions into task-and-motion representations and corrects translation errors~\cite{chen2024autotamp}; VoxPoser composes spatial value maps for model-based planning~\cite{huang2023voxposer}; and Agentic Fast-Slow Planning maps LLM directives to search costs and MPC tracking while refining planner settings with feedback and memory~\cite{chen2026agentic}.

Feedback and memory provide complementary ways to revise language-generated decisions. Inner Monologue feeds execution feedback into subsequent planning~\cite{huang2023inner}; Reflexion stores verbal reflections without updating model parameters~\cite{shinn2023reflexion}; and DiLu retrieves driving experiences to revise unsuccessful decisions, using a memory seeded with manually designed scenarios~\cite{wen2024dilu}. For narrow-space parking, \method{} centers the loop on maneuver-level reasoning: the LLM proposes sparse pose anchors and motion modes, while sequential OCPs and full-footprint validation determine feasibility. Failed attempts trigger bounded online replanning, and accumulated failures drive offline updates of an initially empty parking knowledge base. A candidate update is retained only when the triggering scene and all previously successful scenes pass the same trajectory-acceptance pipeline.

\section{METHODS}

 \method{} comprises an online hierarchical pipeline
(Fig.~\ref{fig:pipeline}) and an offline knowledge self-evolution
loop (Fig.~\ref{fig:knowledge_evolution}). Online, LLM maneuver
decisions are grounded into sequential OCPs, and each trial uses
a fixed knowledge‑base version; offline, accumulated failures
drive candidate knowledge updates that are verified through the
same trajectory‑acceptance pipeline.

\subsection{Online Hierarchical Pipeline}

The online pipeline has three stages: high‑level maneuver
planning, low‑level trajectory optimization, and
feedback‑conditioned replanning. The LLM specifies pose anchors
and motion modes that divide the parking task into segments;
these decisions define terminal conditions for sequential OCPs,
and a validated trajectory is sent to the controller. Each task
allows three planning attempts.

\subsubsection{High-level Maneuver Planning}

A parking scene is defined by a bounded workspace
$\mathcal W\subset\mathbb R^2$, the union of polygonal obstacle
regions $\mathcal O$, an initial pose
$\mathbf p_0=(x_0,y_0,\theta_0)$, and a goal pose $\mathbf p_g$.
The region $\mathcal W\setminus\mathcal O$ is available for vehicle
motion. Let $\mathcal F(\mathbf p)$ denote the oriented rectangular
vehicle footprint at pose $\mathbf p=(x,y,\theta)$. The
collision-free configuration set is
\begin{equation}
\mathcal C_{\mathrm{free}}
=
\left\{
\mathbf p
\;\middle|\;
\mathcal F(\mathbf p)
\subseteq
\mathcal W\setminus\mathcal O
\right\}.
\label{eq:cfree}
\end{equation}
The task is to reach $\mathbf p_g$ from $\mathbf p_0$ through
configurations in $\mathcal C_{\mathrm{free}}$ while satisfying
the vehicle  kinematics and actuation limits.

The decision-making LLM receives the task specification, scene
image, metric geometry, start and goal poses, and vehicle
dimensions. Relevant knowledge is retrieved by matching the
current scene against the accumulated entries, each of which
records a scene description together with explicitly allowed
and forbidden operations. The retrieved entries guide the
selection of staging poses and motion modes. The
LLM generates an ordered maneuver plan with $N$ segments:
\begin{equation}
\Pi
=
\left\{(\mathbf p_i,m_i,e_i)\right\}_{i=1}^{N},
\label{eq:plan}
\end{equation}
where $\mathbf p_i$ is the terminal-pose anchor of segment $i$,
$m_i\in\{\texttt{forward},\texttt{reverse}\}$ specifies its
direction of travel, and $e_i$ is a concise natural-language
rationale explaining why the segment moves this way. Each
rationale is carried through grounding and optimization; when a
segment subsequently fails to solve or validate, its rationale
is returned to the LLM together with the failure evidence, so
that revisions can build on the original intent of the failed
decision. Intermediate anchors
specify staging poses for reorientation and goal approach, and
the final anchor coincides with the goal, $\mathbf p_N=\mathbf p_g$.
Vehicle parameters, objective weights, horizon lengths, and
collision constraints remain system-defined.

\begin{figure*}[!t]
\centering
\includegraphics[width=0.86\textwidth]{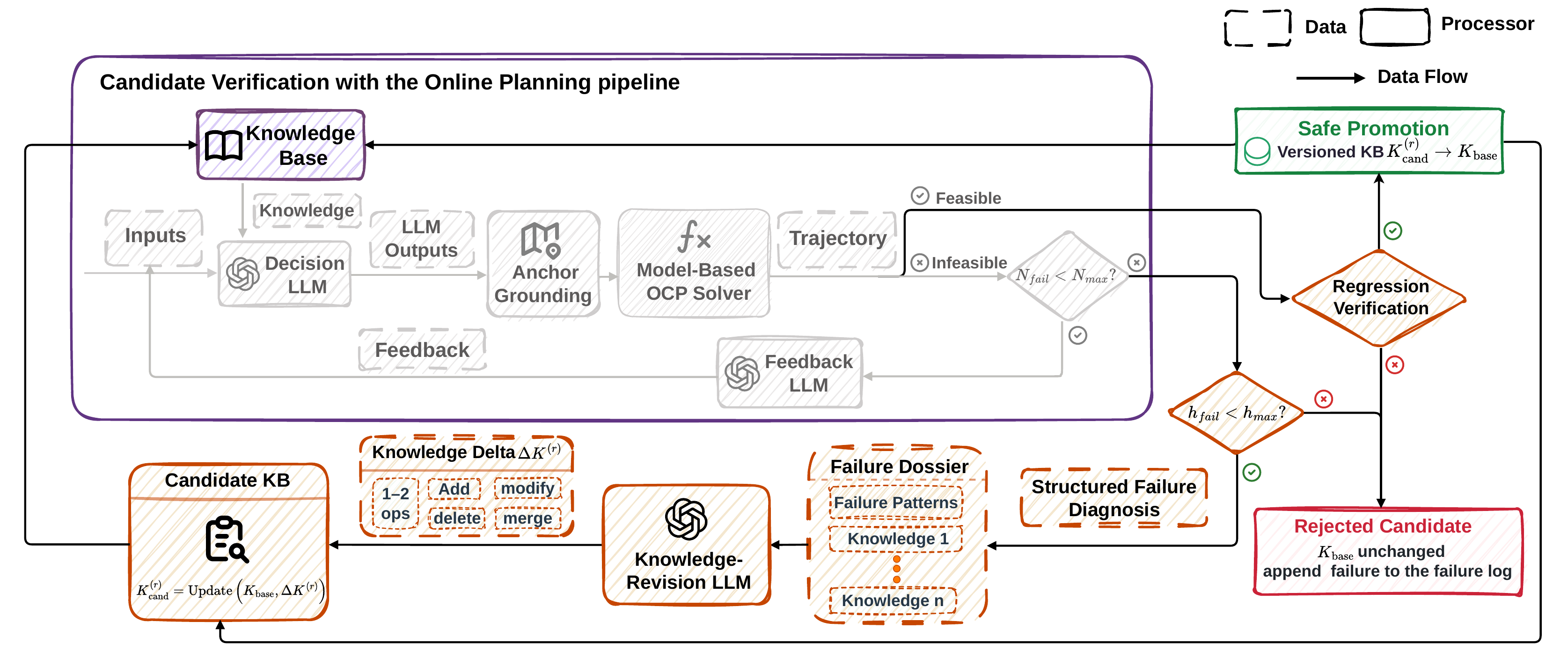}
\caption{Offline knowledge self-evolution loop of \method{}:
a knowledge-revision LLM proposes updates from failure dossiers.
A candidate is retained only if the triggering and all previously
solved scenarios pass the online pipeline's trajectory acceptance
checks. Failed verification supplies evidence for further revision
within the offline budget.}
\label{fig:knowledge_evolution}
\end{figure*}

\subsubsection{Low-level Trajectory Optimization}

The high-level maneuver plan is converted into terminal
constraints and penalties for sequential OCPs via adaptive anchor
grounding. Requiring every intermediate anchor to be reached
exactly can overconstrain these OCPs, even when nearby staging
poses would permit feasible motion. Bounded endpoint adjustments
are therefore allowed, with penalties favoring the nominal
anchors. The final goal remains fixed. We first describe how
the anchors are grounded into terminal regions and penalties,
and then formulate the segment OCPs.

\textbf{Anchor grounding.}
To make the allowed adjustments precise, we attach to each
intermediate anchor $\mathbf p_i$ a positional tolerance $r_i$
and a heading tolerance $\Delta\theta_{\mathrm{tol}}$, which
bound how far the endpoint of segment $i$ may deviate from the
nominal anchor in position and heading, respectively. The
positional tolerance is derived from a clearance radius
$\rho_i$, which measures the vehicle's translational freedom at
the anchor. For an intermediate anchor
$\mathbf p_i=(x_i,y_i,\theta_i)$, $\rho_i$ is the largest radius
within which the vehicle can translate in any direction without
collision while its heading is held at $\theta_i$. With a cap
$\rho_{\mathrm{cap}}$ and planar displacement
$\mathbf d=(d_x,d_y)$, it is defined as
\begin{equation}
\begin{aligned}
\rho_i=\sup\bigl\{&
\rho\in[0,\rho_{\mathrm{cap}}]\;\big|\;
\\
& (x_i+d_x,y_i+d_y,\theta_i)\in\mathcal C_{\mathrm{free}},
\\
& \forall\,\mathbf d\in\mathbb R^2
\text{ with }\|\mathbf d\|_2\leq\rho
\bigr\},\quad i<N.
\end{aligned}
\label{eq:rho}
\end{equation}
The positional tolerance
$r_i$ bounds the positional deviation of the segment endpoint
from its nominal anchor and is computed as
\begin{equation}
r_i
=
\max\left\{
r_{\min},
\min\left(\varepsilon_{\mathrm{task}},k\rho_i\right)
\right\},
\quad i<N.
\label{eq:radius}
\end{equation}
Here, $k$ is a scale factor and $\varepsilon_{\mathrm{task}}$
is the task-level positional tolerance. The lower bound
$r_{\min}$ prevents the tolerance from vanishing.

The terminal pose region $\mathcal T_i$ combines the positional
tolerance $r_i$ with the heading tolerance
$\Delta\theta_{\mathrm{tol}}$. The endpoint of segment $i$ is
required to lie in the set
\begin{equation}
\begin{aligned}
\mathcal T_i=\bigl\{&\mathbf p=(x,y,\theta)\;\big|\;\\
&\left\|(x,y)-(x_i,y_i)\right\|_2\leq r_i,\\
&\left|\operatorname{wrap}(\theta-\theta_i)\right|
 \leq\Delta\theta_{\mathrm{tol}}
\bigr\},
\quad i<N.
\end{aligned}
\label{eq:tset}
\end{equation}
Here, $\operatorname{wrap}$ returns the principal angular
difference. For the final segment, the set reduces to
$\mathcal T_{N}=\{\mathbf p_g\}$.

The terminal region bounds the endpoint adjustment, while a
terminal pose penalty $J_{\mathrm{term}}^i$ favors the nominal
anchor. Its position and heading weights are
$w_{\mathrm{pos}}^i$ and $w_\theta^i$. The position weight is
reduced as the tolerance grows, so that a generous tolerance
weakens the pull of the penalty toward the nominal anchor;
the heading weight, in contrast, is held fixed at
$w_\theta^{\mathrm{int}}$ to keep the approach direction
precise. For intermediate anchors, the position weight is
computed as
\begin{equation}
w_{\mathrm{pos}}(r_i)
=
\max\left(
w_{\mathrm{pos}}^{\min},\;
w_{\mathrm{pos}}^{\max}
\frac{r_{\min}}{\max(r_i,r_{\min})}
\right)
\label{eq:weights}
\end{equation}
where the constants are set empirically to
$w_{\mathrm{pos}}^{\min}=80$, $w_{\mathrm{pos}}^{\max}=200$,
and $w_\theta^{\mathrm{int}}=400$. For the final segment, the
weights are fixed to $w_{\mathrm{pos}}^{\mathrm{goal}}=1000$
and $w_\theta^{\mathrm{goal}}=2000$, likewise set empirically.
The operation $\operatorname{TerminalCost}$ constructs
$J_{\mathrm{term}}^i$ from the anchor and these weights.

The region itself does not certify collision freedom:
$\rho_i$ assumes a fixed heading, whereas $\mathcal T_i$ permits
heading changes and the radius mapping can yield $r_i>\rho_i$.
Collision constraints are therefore imposed independently in
the OCP.

Algorithm~\ref{alg:anchor_soft_constraint} summarizes the mapping
from the maneuver plan $\Pi$ to the grounded plan
$\Gamma$. Intermediate segments receive adaptive
position tolerances and penalty weights; the final segment
retains the exact goal and fixed weights. Each entry $\Gamma_i$
contains the maneuver attributes, positional tolerance,
terminal region, and terminal penalty needed to formulate the
corresponding segment OCP.

\begin{algorithm}[!t]
\caption{Adaptive Soft-Constraint Grounding of LLM Anchors}
\label{alg:anchor_soft_constraint}
\KwInput{Plan $\Pi$; $\mathcal C_{\mathrm{free}}$; goal $\mathbf p_g$;
parameters $(\varepsilon_{\mathrm{task}},k,r_{\min},
\rho_{\mathrm{cap}},\Delta\theta_{\mathrm{tol}})$}
\KwOutput{Grounded plan $\Gamma$}
$N\leftarrow|\Pi|$; $\Gamma\leftarrow\emptyset$\;
\For{$i\leftarrow1$ \KwTo $N$}{
    $(\mathbf p_i,m_i,e_i)\leftarrow\Pi[i]$\;
    \eIf{$i=N$}{
        $\mathbf p_i\leftarrow\mathbf p_g$; $r_i\leftarrow0$;
        $\mathcal T_i\leftarrow\{\mathbf p_g\}$\;
        $(w_{\mathrm{pos}}^i,w_\theta^i)\leftarrow
        (w_{\mathrm{pos}}^{\mathrm{goal}},w_\theta^{\mathrm{goal}})$\;
    }{
        Compute $\rho_i$ using~\eqref{eq:rho}
        and $r_i$ using~\eqref{eq:radius}\;
        Compute $w_{\mathrm{pos}}^i$ using~\eqref{eq:weights}
        and set $w_\theta^i\leftarrow w_\theta^{\mathrm{int}}$\;
        Compute $\mathcal T_i$ using~\eqref{eq:tset}\;
    }
    $J_{\mathrm{term}}^i\leftarrow\operatorname{TerminalCost}
    (\mathbf p_i,w_{\mathrm{pos}}^i,w_\theta^i)$\;
    $\Gamma_i\leftarrow(\mathbf p_i,m_i,e_i,
    r_i,\mathcal T_i,J_{\mathrm{term}}^i)$\;
}
\KwRet $\Gamma=\{\Gamma_i\}_{i=1}^{N}$\;
\end{algorithm}

\textbf{Segment OCP formulation.}
Vehicle motion follows the rear-axle kinematic model with state
$\mathbf x=[x,y,\theta,v,\delta]^\top$ and control
$\mathbf u=[a,\omega_\delta]^\top$, where $v$ is signed longitudinal
velocity, $\delta$ is steering angle, $a$ is longitudinal
acceleration, and $\omega_\delta$ is steering rate. With
wheelbase $L$, the dynamics are

\begin{align}
\dot x &= v\cos\theta, &
\dot y &= v\sin\theta, \nonumber\\
\dot\theta &= \frac{v\tan\delta}{L}, &
\dot v &= a, &
\dot\delta &= \omega_\delta.
\label{eq:dynamics}
\end{align}
For segment $i$, the terminal-state set $H_i$ requires the
terminal pose to lie in $\mathcal T_i$, and the admissible
velocity interval $V(m_i)$ enforces the specified direction of
travel. The terminal cost $\Phi_i$ combines
$J_{\mathrm{term}}^i$ with a penalty that encourages stopping
at the endpoint. Given the initial state $\mathbf x_{\mathrm{init}}$
and the discrete dynamics $f_d$ from~\eqref{eq:dynamics}, each
segment is optimized by solving
\begin{equation}
\begin{aligned}
\min_{\mathbf{x},u}\quad
&
\sum_{t=0}^{T-1}
\ell(\mathbf{x}_t,\mathbf u_t)
+
\Phi_i(\mathbf{x}_T)
\\
\mathrm{s.t.}\quad
&
\mathbf{x}_0=\mathbf{x}_{\mathrm{init}},
\\
&
\mathbf{x}_{t+1}
=
f_d(\mathbf{x}_t,\mathbf u_t),
\\
&
s_{t,c}(\mathbf{x}_t)
\geq
d_{\mathrm{safe}},
\\
&
\mathbf{x}_T\in H_i.
\end{aligned}
\label{eq:segment_ocp}
\end{equation}
Here, $T$ denotes the prediction horizon of the current segment.
The quantity $s_{t,c}$ denotes the full-footprint signed clearance
for constraint $c$ at step $t$, where $c$ indexes obstacle and boundary checks;
$d_{\mathrm{safe}}$ is the required margin.
Control and steering bounds and $v_t\in V(m_i)$ are also imposed.
The stage cost $\ell$ regularizes acceleration, steering rate,
steering angle, velocity, and obstacle slack. Horizon selection
and solver retry settings remain system-defined.

The first OCP starts from the initial vehicle state. Each
subsequent OCP starts from the preceding segment's optimized
terminal state, including velocity and steering angle.
Transferring the full state preserves continuity and allows
the next segment to account for the preceding endpoint
adjustment.

\subsubsection{Feedback and Replanning}

The complete maneuver is checked after all segment OCPs have
been solved. Validation assesses endpoint accuracy, continuity
between segments, satisfaction of the terminal regions, and
footprint feasibility using a sampled vehicle sweep. Trajectories
that pass all feasibility checks are sent to the controller.

When OCP optimization fails or the assembled trajectory fails
validation, solver diagnostics and validation errors are passed
to the feedback LLM together with the rationales of the failed
segments. Its maneuver-level diagnosis guides the decision-making LLM in revising pose anchors and motion modes; the revised plan then follows the same fixed grounding, optimization, and validation procedure.

The online loop terminates when a feasible trajectory is found
or the prescribed attempt budget is exhausted. This feedback
revises the maneuver for the current task; the associated
failure evidence is retained for offline knowledge revision.

\subsection{Offline Knowledge Self-Evolution Loop}

The offline loop in Fig.~\ref{fig:knowledge_evolution} uses accumulated planning failures to revise the decision principles retrieved during high-level planning. The knowledge base is initialized empty (zero-shot), with no hand-authored maneuvers, heuristics, or exemplars. A failure dossier combines observed patterns with relevant entries, and the knowledge-revision LLM proposes a delta of one or two additions, modifications, deletions, or merges. Each generated entry records its applicable scene and explicitly allowed and forbidden operations.

Each candidate knowledge-base version is evaluated through the
online pipeline in Fig.~\ref{fig:pipeline}, with knowledge
retrieved from that version. The resulting maneuver
plans undergo grounding, sequential OCP solution, and trajectory
validation. Online feedback and replanning remain active within
the prescribed attempt budget. The vehicle model, grounding
rules, optimization settings, and acceptance criteria are held
fixed during verification.

A candidate replaces the retained knowledge base only if the triggering scene and all previously successful scenes produce accepted trajectories. Otherwise, the retained version remains unchanged, and each new failure is appended to the dossier for another knowledge revision. The online attempt limit applies to each scene evaluation; a separate offline budget limits candidate updates. The loop ends on promotion or exhaustion of this budget. This regression check protects success on the previously tested scenes under the stated verification protocol.

\section{Experiments}
\label{sec:experiments}

We evaluate \method{} in confined parking environments through
planner comparisons, component ablations, and real-robot
execution. The simulation study examines planning success,
trajectory quality, and closed-loop performance; the physical
experiment demonstrates execution on a differential-drive robot.

\begin{table*}[!t]
\caption{Planning quality and closed-loop performance under a shared Stanley--PID controller.}
\label{tab:planning_execution_compare}
\centering
\begingroup
\small
\setlength{\tabcolsep}{3pt}
\renewcommand{\arraystretch}{1.12}
\begin{tabularx}{\textwidth}{@{}l*{8}{>{\centering\arraybackslash}X}@{}}
\toprule
& \multicolumn{6}{c}{Planning}
& \multicolumn{2}{c}{Closed-loop execution} \\
\cmidrule(lr){2-7}\cmidrule(l){8-9}
Method
& \makecell{Succ.\\(\%)$\uparrow$}
& \makecell{Gear shifts\\$\downarrow$}
& \makecell{Mean $|\kappa|$\\($\mathrm{m}^{-1}$)$\downarrow$}
& \makecell{Max $|d\kappa/ds|$\\($\mathrm{m}^{-2}$)$\downarrow$}
& \makecell{Pos. err.\\(m)}
& \makecell{Orie. err.\\(deg)}
& \makecell{Succ.\\(\%)$\uparrow$}
& \makecell{Coll.\\(\%)} \\
\midrule
HA$^*$
& 92.1 & 2.571 & 0.158 & 4.000
& --- & --- & 55.3 & 40.0 \\
Direct OCP$^\dagger$
& 52.6 & 2.500 & 0.151 & 3.702
& \textbf{0.003} & \textbf{0.068} & 47.4 & \textbf{10.0} \\
HA$^*$+OCP
& 76.3 & 2.138 & 0.158 & 3.556
& 0.028 & 0.529 & 63.2 & 17.2 \\
SMC-OCP$^{\ddagger}$
& 60.5 & 2.261 & 0.162 & 3.776
& 0.029 & 0.482 & 47.4 & 21.7 \\
\rowcolor{oursgreen}
\textbf{\method{}}
& \textbf{94.7} & \textbf{1.593}
& \textbf{0.116} & \textbf{2.437}
& 0.006 &  0.112
& \textbf{84.2} & 11.1 \\
\bottomrule
\end{tabularx}
\par\smallskip
\begin{minipage}{\textwidth}
\footnotesize
Please refer to \textit{Evaluation metrics} for the definitions and
conditioning of each quantity. Dashes indicate that HA$^*$ has no
comparable continuous terminal-pose quantity. $^\dagger$Direct OCP uses
a 2D A$^*$ reference and signed-speed optimization; its planning-quality
statistics are computed over validated plans, while its collision
statistics are conditioned on validated plans that enter execution.
$^\ddagger$SMC-OCP uses fixed sparse anchors extracted from a
Hybrid A$^*$ route; see Experimental Setup for its construction. Its
planning-quality values are averaged over validated plans, while collision
values use the validated plans that proceed to execution.
\end{minipage}
\endgroup
\end{table*}

\begin{figure*}[!t]
\centering
\includegraphics[width=\textwidth]{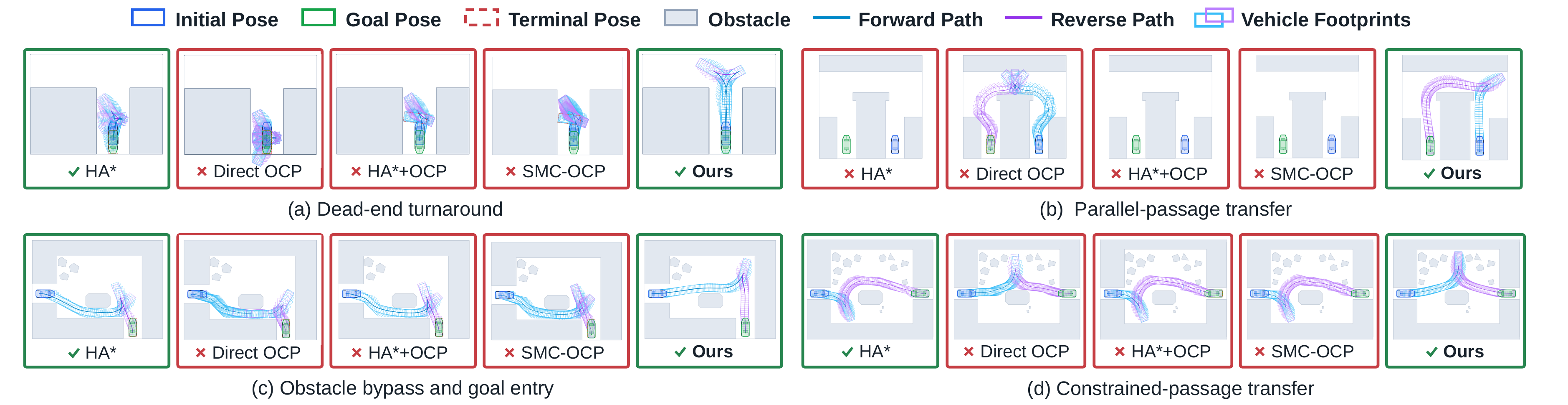}
\caption{Qualitative results across four representative confined parking
scenarios. Each group shows HA*, Direct OCP, HA*+OCP, and \method{}
(Ours) from left to right. Green frames with check marks indicate
success, while red frames with crosses indicate failure.}
\label{fig:qualitative_compare}
\end{figure*}

\subsection{Experimental Setup}

\textbf{Scenes and vehicle.}
The simulation benchmark comprises parking tasks in confined environments with
nonconvex free space. Each task specifies an initial
pose and a goal pose. The car-like model has a
$4.50\times1.85$~m footprint, a $2.70$~m wheelbase, and a
minimum rear-axle turning radius of $5.00$~m.

\textbf{Controller and execution.}
Each validated reference is tracked by the same discrete Stanley--PID
controller at $0.05$~s. The cruise-speed cap is $0.30$~m/s,  the Stanley gain is $0.8$, and the PID
gains are $(K_p,K_i,K_d)=(1.5,0.1,0.03)$. Acceleration is saturated to
$[-0.64,0.64]$~m/s$^2$ and steering-rate magnitude to $0.70$~rad/s.
These settings and the execution model are shared across
planners; curvature, steering-rate, and braking limits can further
reduce speed.

\textbf{Compared planners.}
We compare five planners under a common vehicle model, OCP solver,
validator, and controller. The reference set spans four paradigms.
HA$^*$ is the search-only nonholonomic reference. Direct OCP optimizes a 2-D A$^*$ reference
with signed speed, whereas HA$^*$+OCP refines the dense Hybrid A$^*$
reference with the shared OCP. SMC-OCP (Sparse Maneuver-Corridor OCP) is a deterministic, non-LLM
baseline. It uses the same Hybrid A$^*$ search only to extract four
fixed arc-length anchors from each gear-consistent segment; it then
re-optimizes these sparse references without receiving the dense
search trajectory or a state warm start. In contrast, \method{}
selects staging poses, motion modes, and segment order before solving
sequential OCPs.

\textbf{Evaluation metrics.}
Planning success requires the generated reference trajectory to pass
the common trajectory validator, including continuity, terminal
conditions, collision checks, and boundary checks. Closed-loop success
requires successful planning followed by execution to within $0.50$~m
of the goal position and $10^\circ$ of the goal heading, without
obstacle collision or departure from the drivable region. Both success
rates use the corresponding evaluation runs. Collision rate is
conditional on runs that pass planning validation and proceed to
controller execution, including unsuccessful executions; a run is
counted once if an obstacle collision or boundary violation occurs.
Planning failures are excluded from this conditional statistic.

Planning-quality metrics are averaged over validated runs. They
comprise gear shifts, mean absolute curvature $|\kappa|$
($\mathrm{m}^{-1}$), and the mean per-trajectory maximum absolute
spatial curvature gradient $|d\kappa/ds|$ ($\mathrm{m}^{-2}$), where $s$
denotes arc length. Terminal position and orientation errors measure
the final pose error of the planned trajectory and are distinct from
errors after controller execution. They are reported for methods with
a comparable validated terminal pose; HA$^*$ has no such continuous
terminal-pose quantity.

\textbf{Evaluation protocol and planning budget.}
All configurations use the same scenes, vehicle, controller, validator,
and metric definitions. All LLM components use gpt-5.6-sol. Under
fixed scenes and solver settings, the non-LLM search and optimization
baselines have negligible run-to-run stochasticity and are therefore
evaluated once per scene. Because LLM sampling can change the proposed
maneuver structure, \method{} and each LLM-based ablation are evaluated
in three independent runs per scene, with no best-run selection. For
\method{} and the LLM-based ablations, each run allows at most three
planning attempts; the deterministic baselines use their fixed solver
output. When feedback is enabled, later proposals are conditioned on
solver and validation failures, while the no-feedback variant retains
the same attempt limit without those diagnostics. The knowledge-base version is fixed throughout each run,
and every proposal is grounded, optimized, and validated.

\subsection{Experiment 1: Planner  and Closed-Loop Performance}

This experiment asks whether maneuver-level staging and motion-mode
decisions improve both trajectory feasibility and closed-loop execution
beyond search-only planning and search-conditioned OCP refinement.
Table~\ref{tab:planning_execution_compare} compares \method{} with
four reference strategies under the shared controller.

Table~\ref{tab:planning_execution_compare} shows that \method{}
achieves the highest planning and closed-loop success rates
($94.7\%$ and $84.2\%$). HA$^*$ has similar planning success
($92.1\%$) but lower closed-loop success; HA$^*$+OCP improves execution
success over HA$^*$ despite lower planning coverage. SMC-OCP and Direct
OCP achieve lower planning and closed-loop success under the same
evaluation criteria. Among validated runs, \method{} also uses fewer
gear shifts and has lower mean curvature and mean peak
$|d\kappa/ds|$.

Direct OCP reports slightly smaller terminal errors
($0.003$~m/$0.068^\circ$ versus $0.006$~m/$0.112^\circ$) and a
slightly lower aggregate conditional collision rate ($10.0\%$ versus
$11.1\%$). Terminal-pose errors are averaged over each method's
validated plans, whereas collision rates are conditioned on the subset
that proceeds to execution. The resulting Direct OCP subset can therefore
favor comparatively easy scenes. A scene-matched diagnostic on the same 20 scene instances that Direct OCP passes to execution shows that SE-LLM–OCP collides in 2 of 60 runs ($3.33\%$), versus $10.0\%$ (2/20) for Direct OCP. Because this subset is defined by Direct OCP's observed
outcomes, the diagnostic is descriptive and does not replace the
full-benchmark rates in Table~\ref{tab:planning_execution_compare}; the
lower aggregate Direct OCP rate should not be interpreted as uniformly
safer behavior on the same scenes.

Figure~\ref{fig:qualitative_compare} gives a qualitative mechanism
consistent with the aggregate results. In the representative cases,
\method{} places major heading changes in open areas and reserves the
confined goal region for final entry, whereas the reference-refinement
baselines retain search-selected staging choices. The figure illustrates
this possible mechanism and is not intended as an exhaustive comparison.

\subsection{Experiment 2: Ablation Study}

We examine the evolved knowledge base, failure feedback, and
intermediate soft constraints by removing each component in turn.
The scene inputs, LLM, vehicle model, controller, and remaining
planning settings are held fixed. Table~\ref{tab:component_ablation}
reports the results.

\begin{table}[!ht]
\caption{Component ablations under shared evaluation settings.}
\label{tab:component_ablation}
\centering
\begingroup
\footnotesize
\setlength{\tabcolsep}{3pt}
\renewcommand{\arraystretch}{1.15}
\begin{tabularx}{\columnwidth}{@{}l*{3}{>{\centering\arraybackslash}X}@{}}
\toprule
& Planning & \multicolumn{2}{c}{Closed-loop execution} \\
\cmidrule(lr){2-2}\cmidrule(l){3-4}
Configuration & \makecell{Succ.\\(\%)$\uparrow$}
& \makecell{Succ.\\(\%)$\uparrow$}
& \makecell{Coll.\\(\%)$\downarrow$} \\
\midrule
w/o evolved knowledge & 85.1 & 74.6 & 12.4 \\
w/o failure feedback & 79.8 & 68.4 & 13.2 \\
w/o soft constraints
& 83.3 & 76.3 & \textbf{8.4} \\
\rowcolor{oursgreen}
Full method & \textbf{94.7} & \textbf{84.2} & 11.1 \\
\bottomrule
\end{tabularx}
\par\smallskip
\begin{minipage}{\columnwidth}
\footnotesize
Success rates use all runs; collision rates count an obstacle collision
or boundary violation once per run, over successfully planned runs that
enter execution (including execution failures).
\end{minipage}
\endgroup
\end{table}

\textbf{Evolved knowledge.}
The knowledge base is initialized empty and populated entirely
through automated offline self-evolution. The variant without
evolved knowledge therefore uses an empty knowledge base. In the full method,
the accumulated knowledge guides staging-pose selection,
approach headings, and motion modes before OCP construction.
Using the evolved knowledge increases planning success from
$85.1\%$ to $94.7\%$ and closed-loop success from $74.6\%$
to $84.2\%$. These gains show the benefit of transferring
accumulated planning experience to online maneuver selection.

\textbf{Failure feedback.}
This variant retains the same computational budget and
planning-attempt limit as the full method, but subsequent
proposals receive no feedback from earlier failures.
The full method uses solver and validation feedback to revise
pose anchors and motion modes in response to the difficulties
encountered during previous attempts. With the same budget,
failure feedback increases planning success from $79.8\%$
to $94.7\%$ and closed-loop success from $68.4\%$ to $84.2\%$.
The improvement shows the value of using failed attempts to
guide subsequent maneuver proposals.

\begin{figure}[!t]
\centering
\includegraphics[width=\columnwidth,trim=20bp 12bp 38bp 12bp,clip]{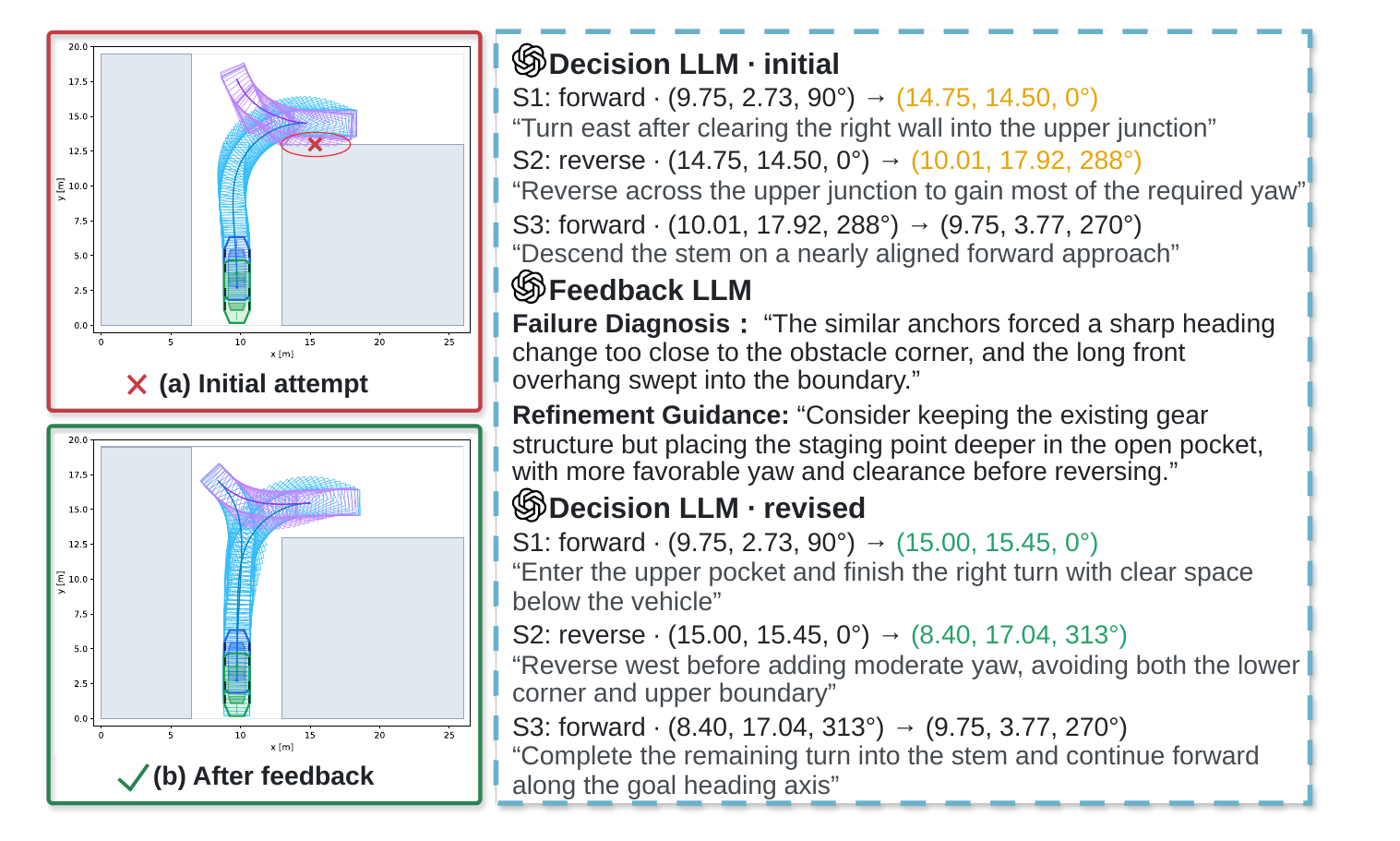}
\caption{Failure-conditioned recovery through anchor revision.
(a) The initial reverse segment fails collision checks.
(b) Revised
anchors produce a validated trajectory with the same
forward--reverse--forward sequence. Right: LLM decisions,
segment rationales, and selected feedback excerpts. Poses are
$(x,y,\theta)$ in metres and degrees; headings are rounded
for display.}
\label{fig:feedback_case}
\end{figure}

\Needspace{5\baselineskip}
Figure~\ref{fig:feedback_case} illustrates feedback-conditioned
maneuver revision. The initial attempt solves the forward
staging segment, but the reverse segment fails with a footprint
collision. The feedback attributes the collision to the front
overhang sweeping into an obstacle corner. The revised decision
moves the staging anchor deeper into the open pocket and adjusts
the reverse endpoint, redistributing the heading change across
the remaining segments. The resulting trajectory passes
validation with the same forward--reverse--forward sequence.

\textbf{Intermediate soft constraints.}
This variant solves each segment as a separate start-to-goal
problem, applying the task-terminal constraints at every segment
endpoint. Treating each intermediate anchor as a task goal
restricts where adjacent segments can connect, even when a nearby
pose could serve the same maneuver. The full method allows
intermediate poses to adjust during grounding while retaining
the final goal constraints. This increases planning success
from $83.3\%$ to $94.7\%$ and closed-loop success from $76.3\%$
to $84.2\%$, supporting the use of soft constraints to accommodate
approximate LLM anchors. The ablated variant has a lower collision
rate among executed plans ($8.4\%$ versus $11.1\%$), while the
full method completes more tasks overall.

\subsection{Experiment 3: Real-Robot Experiment}

\begin{figure}[!t]
\centering
\begin{minipage}[b]{0.46\columnwidth}
\centering
\includegraphics[width=\linewidth,height=0.80in,keepaspectratio]{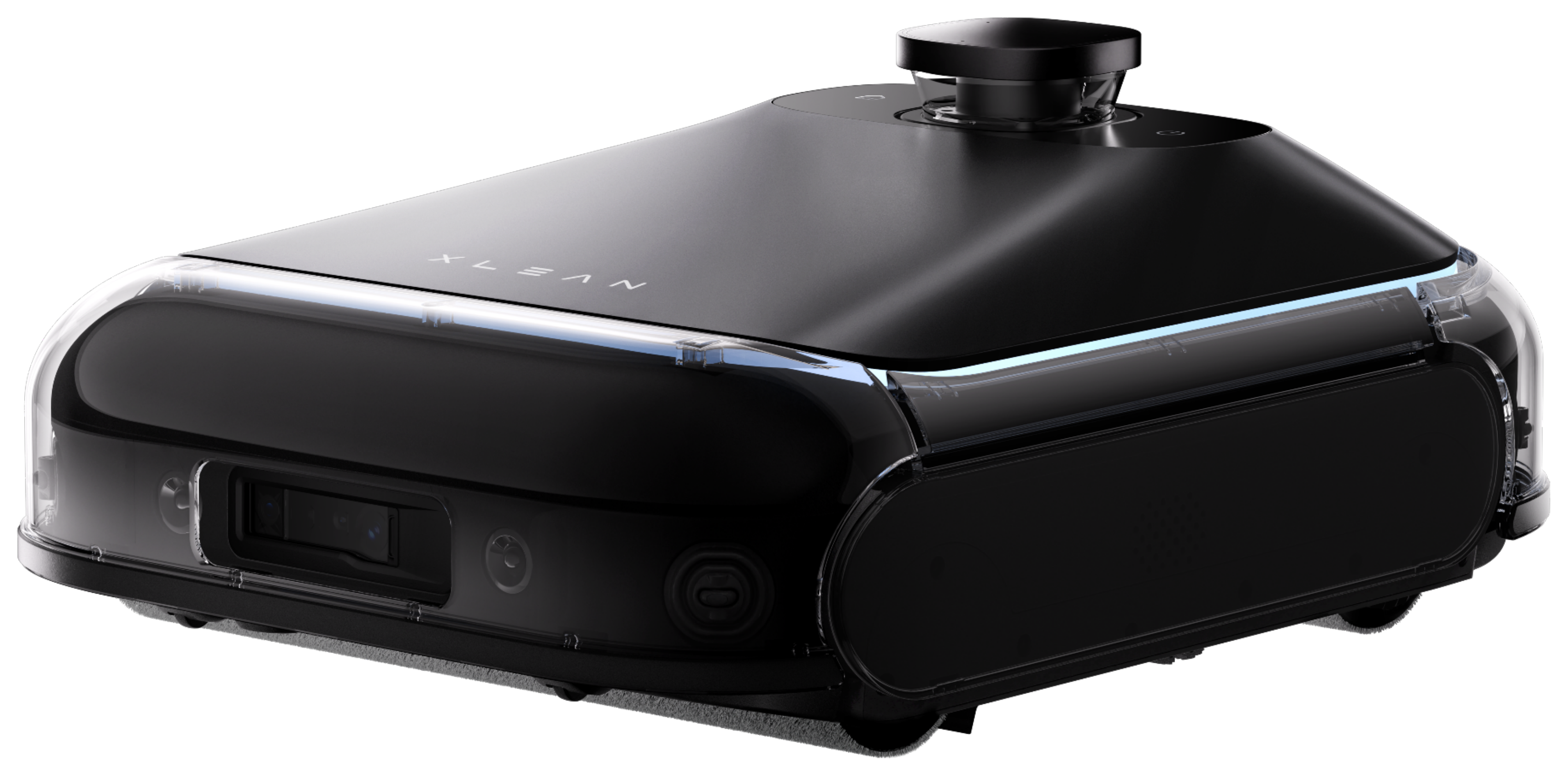}
\par\smallskip
{\footnotesize (a) Differential-drive platform}
\end{minipage}\hfill
\begin{minipage}[b]{0.46\columnwidth}
\centering
\includegraphics[width=\linewidth,height=0.80in,keepaspectratio]{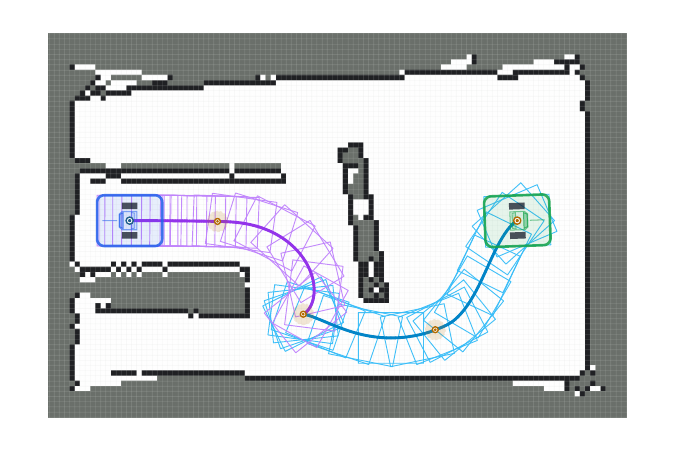}
\par\smallskip
{\footnotesize (b) Simulated reference}
\end{minipage}

\par\smallskip
\begin{minipage}[c]{\columnwidth}
\centering
\includegraphics[width=\linewidth]{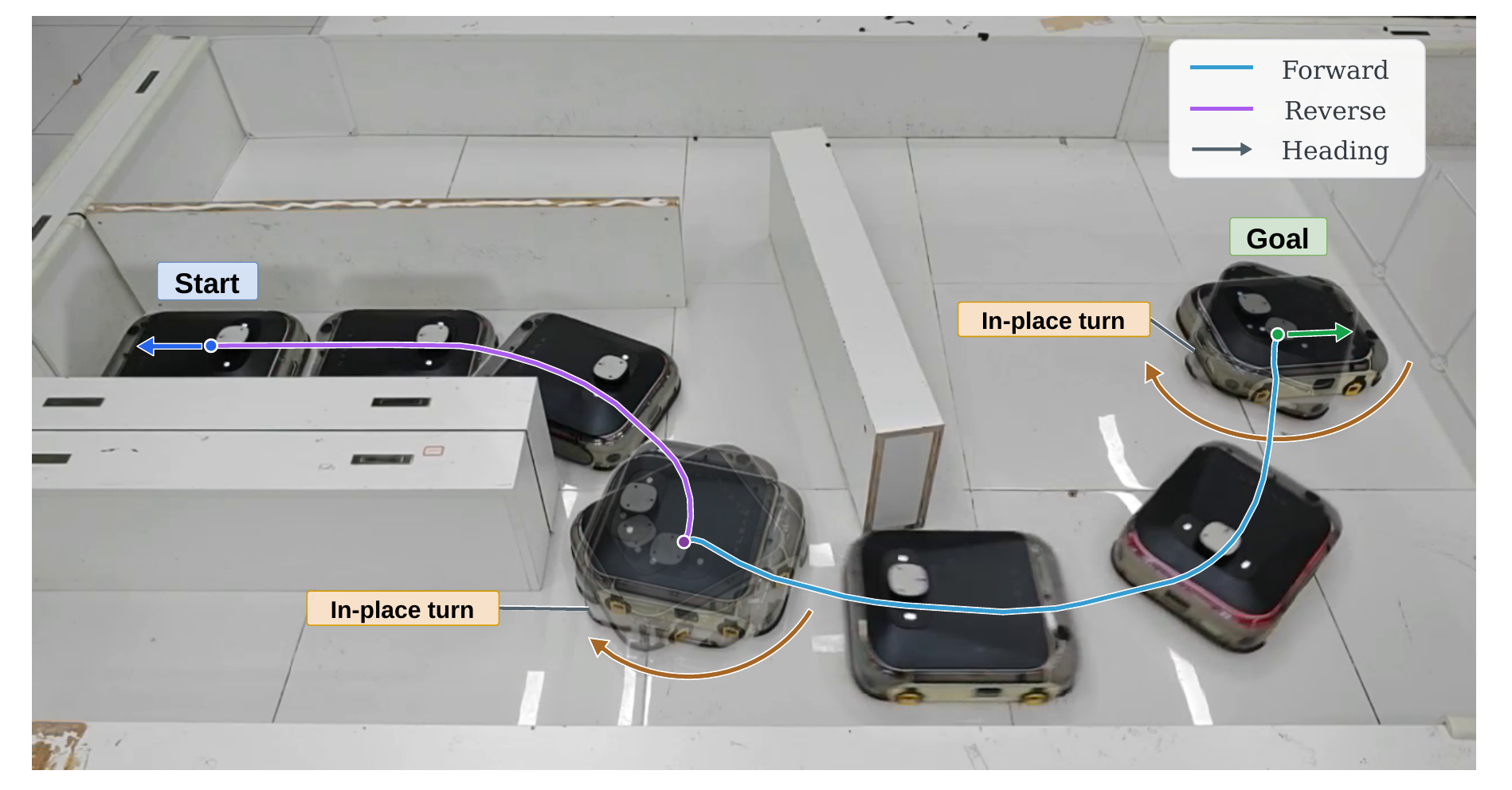}
\par\smallskip
{\footnotesize (c) Real-robot execution}
\end{minipage}
\caption{Real-robot experiment and corresponding simulated reference.
(a) Differential-drive robot equipped with a 2D LiDAR. (b) Simulated
reference with intermediate pose markers and sampled footprints.
(c) Successive robot poses overlaid to show the executed motion.
Purple and blue curves indicate reverse and forward motion,
respectively; the curves in (b) are the reference, those in (c)
the motion sequence.}
\label{fig:real_robot}
\end{figure}

We conducted a physical experiment with the differential-drive
robot shown in Fig.~\ref{fig:real_robot}(a). The platform is
equipped with a 2D LiDAR. The task requires the robot to leave a
narrow aisle, pass around a partition, and reach the marked
goal in a confined indoor environment.

Figure~\ref{fig:real_robot}(b) shows the corresponding simulated
reference, produced by the same planning pipeline with the
segment OCPs re-solved under differential-drive kinematics
instead of the car-like model. A reverse segment brings the
robot out of the starting
aisle into the connecting area, followed by a forward segment
around the partition toward the goal. The intermediate pose
markers indicate the staged maneuver, and the sampled footprints
show the robot configurations along the reference.

The physical execution in Fig.~\ref{fig:real_robot}(c) follows
the same reverse-then-forward maneuver structure. The robot
reverses out of the aisle, reorients in the connecting area,
and moves forward around the partition. In-place rotations are
visible at the intermediate configuration and at the goal.
The first separates departure from the confined aisle from
the forward approach; the second adjusts the heading at the
marked goal. The overlaid poses show the executed motion,
while the colored curves illustrate its forward and reverse
portions.

This experiment provides qualitative evidence that the reverse
departure and forward goal approach can be carried through to
physical execution. The intermediate staging configuration
provides room to reorient, while in-place rotations separate
translation from heading adjustment.

\par\vspace{6pt}\noindent

\section{Conclusion}

We presented \method{}, a parking framework that bridges LLM-based maneuver reasoning and model-based trajectory optimization. The framework uses a knowledge base evolved from scratch to guide staging poses and motion modes, grounds approximate intermediate anchors through adaptive soft constraints, and converts solver and validation failures into feedback for bounded replanning. Simulation comparisons and ablations demonstrated improved planning and closed-loop performance in narrow, nonconvex environments, while a differential-drive robot experiment provided qualitative evidence of physical execution. Future work will focus on repeated quantitative real-robot evaluation and extension to additional kinematic platforms and parking environments.

\bibliographystyle{IEEEtran}
\bibliography{references}

@IEEEtranBSTCTL{BSTcontrol,
  CTLuse_forced_etal       = {yes},
  CTLmax_names_forced_etal = {6},
  CTLnames_show_etal       = {1},
  CTLdash_repeated_names  = {no}
}

@inproceedings{dolgov2008practical,
  author     = {Dolgov, Dmitri and Thrun, Sebastian and Montemerlo, Michael and Diebel, James},
  title      = {Practical search techniques in path planning for autonomous driving},
  booktitle  = {Proc. 1st Int. Symp. Search Tech. Artif. Intell. Robot. (STAIR)},
  year       = {2008},
}

@inproceedings{ratliff2009chomp,
  author     = {Ratliff, Nathan and Zucker, Matthew and Bagnell, J. Andrew and Srinivasa, Siddhartha},
  title      = {{CHOMP}: Gradient optimization techniques for efficient motion planning},
  booktitle  = {Proc. IEEE Int. Conf. Robot. Automat. (ICRA)},
  year       = {2009},
  pages      = {489--494},
  doi        = {10.1109/ROBOT.2009.5152817},
}

@article{zhang2021optimization,
  author     = {Zhang, Xiaojing and Liniger, Alexander and Borrelli, Francesco},
  title      = {Optimization-based collision avoidance},
  journal    = {IEEE Trans. Control Syst. Technol.},
  volume     = {29},
  number     = {3},
  pages      = {972--983},
  month      = may,
  year       = {2021},
  doi        = {10.1109/TCST.2019.2949540},
}

@inproceedings{ichter2023saycan,
  author     = {Ichter, Brian and Brohan, Anthony and Chebotar, Yevgen and Finn, Chelsea and Hausman, Karol and Herzog, Alexander and Ho, Daniel and Ibarz, Julian and Irpan, Alex and Jang, Eric and Julian, Ryan and Kalashnikov, Dmitry and Levine, Sergey and Lu, Yao and Parada, Carolina and Rao, Kanishka and Sermanet, Pierre and Toshev, Alexander T. and Vanhoucke, Vincent and Xia, Fei and Xiao, Ted and Xu, Peng and Yan, Mengyuan and Brown, Noah and Ahn, Michael and Cortes, Omar and Sievers, Nicolas and Tan, Clayton and Xu, Sichun and Reyes, Diego and Rettinghouse, Jarek and Quiambao, Jornell and Pastor, Peter and Luu, Linda and Lee, Kuang-Huei and Kuang, Yuheng and Jesmonth, Sally and Joshi, Nikhil J. and Jeffrey, Kyle and Ruano, Rosario Jauregui and Hsu, Jasmine and Gopalakrishnan, Keerthana and David, Byron and Zeng, Andy and Fu, Chuyuan Kelly},
  title      = {Do as {I} can, not as {I} say: Grounding language in robotic affordances},
  booktitle  = {Proc. 6th Conf. Robot Learn. (CoRL)},
  series     = {Proc. Mach. Learn. Res.},
  volume     = {205},
  year       = {2023},
  pages      = {287--318},
}

@inproceedings{liang2023code,
  author     = {Liang, Jacky and Huang, Wenlong and Xia, Fei and Xu, Peng and Hausman, Karol and Ichter, Brian and Florence, Pete and Zeng, Andy},
  title      = {Code as policies: Language model programs for embodied control},
  booktitle  = {Proc. IEEE Int. Conf. Robot. Automat. (ICRA)},
  year       = {2023},
  pages      = {9493--9500},
  doi        = {10.1109/ICRA48891.2023.10160591},
}

@inproceedings{rana2023sayplan,
  author     = {Rana, Krishan and Haviland, Jesse and Garg, Sourav and Abou-Chakra, Jad and Reid, Ian and S{\"u}nderhauf, Niko},
  title      = {{SayPlan}: Grounding large language models using {3D} scene graphs for scalable robot task planning},
  booktitle  = {Proc. 7th Conf. Robot Learn. (CoRL)},
  series     = {Proc. Mach. Learn. Res.},
  volume     = {229},
  year       = {2023},
  pages      = {23--72},
}

@misc{mao2023gptdriver,
  author     = {Mao, Jiageng and Qian, Yuxi and Ye, Junjie and Zhao, Hang and Wang, Yue},
  title      = {{GPT-Driver}: Learning to drive with {GPT}},
  year       = {2023},
  note       = {arXiv:2310.01415},
  doi        = {10.48550/arXiv.2310.01415},
}

@inproceedings{tian2025drivevlm,
  author     = {Tian, Xiaoyu and Gu, Junru and Li, Bailin and Liu, Yicheng and Wang, Yang and Zhao, Zhiyong and Zhan, Kun and Jia, Peng and Lang, XianPeng and Zhao, Hang},
  title      = {{DriveVLM}: The convergence of autonomous driving and large vision-language models},
  booktitle  = {Proc. 8th Conf. Robot Learn. (CoRL)},
  series     = {Proc. Mach. Learn. Res.},
  volume     = {270},
  year       = {2025},
  pages      = {4698--4726},
}

@article{reeds1990optimal,
  author     = {Reeds, James A. and Shepp, Lawrence A.},
  title      = {Optimal paths for a car that goes both forwards and backwards},
  journal    = {Pacific J. Math.},
  volume     = {145},
  number     = {2},
  pages      = {367--393},
  month      = oct,
  year       = {1990},
  doi        = {10.2140/pjm.1990.145.367},
}

@article{li2022lightweight,
  author     = {Li, Bai and Acarman, Tankut and Zhang, Youmin and Ouyang, Yakun and Yaman, Cagdas and Kong, Qi and Zhong, Xiang and Peng, Xiaoyan},
  title      = {Optimization-based trajectory planning for autonomous parking with irregularly placed obstacles: A lightweight iterative framework},
  journal    = {IEEE Trans. Intell. Transp. Syst.},
  volume     = {23},
  number     = {8},
  pages      = {11970--11981},
  month      = aug,
  year       = {2022},
  doi        = {10.1109/TITS.2021.3109011},
}

@article{han2024spatiotemporal,
  author     = {Han, Zhichao and Wu, Yuwei and Li, Tong and Zhang, Lu and Pei, Liuao and Xu, Long and Li, Chengyang and Ma, Changjia and Xu, Chao and Shen, Shaojie and Gao, Fei},
  title      = {An efficient spatial-temporal trajectory planner for autonomous vehicles in unstructured environments},
  journal    = {IEEE Trans. Intell. Transp. Syst.},
  volume     = {25},
  number     = {2},
  pages      = {1797--1814},
  month      = feb,
  year       = {2024},
  doi        = {10.1109/TITS.2023.3315320},
}

@article{han2023rda,
  author     = {Han, Ruihua and Wang, Shuai and Wang, Shuaijun and Zhang, Zeqing and Zhang, Qianru and Eldar, Yonina C. and Hao, Qi and Pan, Jia},
  title      = {{RDA}: An accelerated collision free motion planner for autonomous navigation in cluttered environments},
  journal    = {IEEE Robot. Autom. Lett.},
  volume     = {8},
  number     = {3},
  pages      = {1715--1722},
  month      = mar,
  year       = {2023},
  doi        = {10.1109/LRA.2023.3242138},
}

@inproceedings{yang2024e2eparking,
  author     = {Yang, Yunfan and Chen, Denglong and Qin, Tong and Mu, Xiangru and Xu, Chunjing and Yang, Ming},
  title      = {{E2E Parking}: Autonomous parking by the end-to-end neural network on the {CARLA} simulator},
  booktitle  = {Proc. IEEE Intell. Veh. Symp. (IV)},
  year       = {2024},
  pages      = {2375--2382},
  doi        = {10.1109/IV55156.2024.10588551},
}

@inproceedings{li2024parkinge2e,
  author     = {Li, Changze and Ji, Ziheng and Chen, Zhe and Qin, Tong and Yang, Ming},
  title      = {{ParkingE2E}: Camera-based end-to-end parking network, from images to planning},
  booktitle  = {Proc. IEEE/RSJ Int. Conf. Intell. Robots Syst. (IROS)},
  year       = {2024},
  pages      = {13206--13212},
  doi        = {10.1109/IROS58592.2024.10801763},
}

@misc{zheng2026multipark,
  author     = {Zheng, Han and Zhou, Zikang and Zhang, Guli and Wang, Zhepei and Wang, Kaixuan and Li, Peiliang and Shen, Shaojie and Yang, Ming and Qin, Tong},
  title      = {{MultiPark}: Multimodal parking transformer with next-segment prediction},
  year       = {2025},
  note       = {arXiv:2508.11537},
  doi        = {10.48550/arXiv.2508.11537},
}

@inproceedings{huang2022zeroshot,
  author     = {Huang, Wenlong and Abbeel, Pieter and Pathak, Deepak and Mordatch, Igor},
  title      = {Language models as zero-shot planners: Extracting actionable knowledge for embodied agents},
  booktitle  = {Proc. 39th Int. Conf. Mach. Learn. (ICML)},
  series     = {Proc. Mach. Learn. Res.},
  volume     = {162},
  year       = {2022},
  pages      = {9118--9147},
}

@inproceedings{chen2024autotamp,
  author     = {Chen, Yongchao and Arkin, Jacob and Dawson, Charles and Zhang, Yang and Roy, Nicholas and Fan, Chuchu},
  title      = {{AutoTAMP}: Autoregressive task and motion planning with {LLM}s as translators and checkers},
  booktitle  = {Proc. IEEE Int. Conf. Robot. Automat. (ICRA)},
  year       = {2024},
  pages      = {6695--6702},
  doi        = {10.1109/ICRA57147.2024.10611163},
}

@inproceedings{huang2023voxposer,
  author     = {Huang, Wenlong and Wang, Chen and Zhang, Ruohan and Li, Yunzhu and Wu, Jiajun and Fei-Fei, Li},
  title      = {{VoxPoser}: Composable {3D} value maps for robotic manipulation with language models},
  booktitle  = {Proc. 7th Conf. Robot Learn. (CoRL)},
  series     = {Proc. Mach. Learn. Res.},
  volume     = {229},
  year       = {2023},
  pages      = {540--562},
}

@inproceedings{huang2023inner,
  author     = {Huang, Wenlong and Xia, Fei and Xiao, Ted and Chan, Harris and Liang, Jacky and Florence, Pete and Zeng, Andy and Tompson, Jonathan and Mordatch, Igor and Chebotar, Yevgen and Sermanet, Pierre and Jackson, Tomas and Brown, Noah and Luu, Linda and Levine, Sergey and Hausman, Karol and Ichter, Brian},
  title      = {Inner monologue: Embodied reasoning through planning with language models},
  booktitle  = {Proc. 6th Conf. Robot Learn. (CoRL)},
  series     = {Proc. Mach. Learn. Res.},
  volume     = {205},
  year       = {2023},
  pages      = {1769--1782},
}

@inproceedings{shinn2023reflexion,
  author     = {Shinn, Noah and Cassano, Federico and Gopinath, Ashwin and Narasimhan, Karthik and Yao, Shunyu},
  title      = {{Reflexion}: Language agents with verbal reinforcement learning},
  booktitle  = {Adv. Neural Inf. Process. Syst. (NeurIPS)},
  volume     = {36},
  year       = {2023},
  pages      = {8634--8652},
  doi        = {10.52202/075280-0377},
}

@inproceedings{wen2024dilu,
  author     = {Wen, Licheng and Fu, Daocheng and Li, Xin and Cai, Xinyu and Ma, Tao and Cai, Pinlong and Dou, Min and Shi, Botian and He, Liang and Qiao, Yu},
  title      = {{DiLu}: A knowledge-driven approach to autonomous driving with large language models},
  booktitle  = {Proc. 12th Int. Conf. Learn. Represent. (ICLR)},
  year       = {2024},
}

@misc{chen2026agentic,
  author     = {Chen, Jiayi and Wang, Shuai and Zhu, Guangxu and Xu, Chengzhong},
  title      = {Bridging large-model reasoning and real-time control via agentic fast-slow planning},
  year       = {2026},
  note       = {arXiv:2604.01681},
  doi        = {10.48550/arXiv.2604.01681},
}

\end{document}